\documentclass[10pt,twocolumn,letterpaper]{article}

\usepackage{iccv}
\usepackage{times}
\usepackage{epsfig}
\usepackage{graphicx}
\usepackage{amsmath}
\usepackage{amssymb}

\usepackage[pagebackref=true,breaklinks=true,letterpaper=true,colorlinks,bookmarks=false]{hyperref}

\usepackage[capitalize]{cleveref}
\crefname{section}{Sec.}{Secs.}
\Crefname{section}{Section}{Sections}
\Crefname{table}{Table}{Tables}
\crefname{table}{Tab.}{Tabs.}

\usepackage{url}            
\usepackage{amsfonts}       
\usepackage{nicefrac}       
\usepackage{microtype}      
\usepackage{xcolor}         
\usepackage{times}
\usepackage{epsfig}
\usepackage{makecell}

\usepackage[accsupp]{axessibility}
\usepackage{enumerate}
\usepackage{enumitem}
\usepackage{diagbox}
\usepackage{bbding}
\usepackage{multirow}
\usepackage{threeparttable}
\usepackage{booktabs}

\def\bv #1{\boldsymbol{\rm{#1}}}

\def\onedot{.}
\def\eg{\textit{e.g}\onedot} 
\def\ie{\textit{i.e}\onedot} 
 
\def\etc{{etc}\onedot} 
 
\def\etal{\textit{et al}\onedot~}

\def\cama{\mathrm{A}}
\def\camb{\mathrm{B}}
\def\c{\mathrm{c}}
\def\h{\mathrm{h}}
\def\rec{\mathrm{rec}}

\newcommand{\Tref}[1]{Tab.~\ref{#1}}

\newcommand{\eref}[1]{Eq.~(\ref{#1})}
\newcommand{\fref}[1]{Fig.~\ref{#1}}
\newcommand{\sref}[1]{Sec.~\ref{#1}}

\definecolor{ssred}{rgb}{0.8,0.0,0.0}
\definecolor{sgreen}{rgb}{0.2,0.6,0.15}
\definecolor{sblue}{rgb}{0,0.3,0.9}

\definecolor{light-gray}{gray}{0.92}
\definecolor{dark-gray}{gray}{0.52}
\definecolor{sred}{gray}{0.32}

\iccvfinalcopy 

\def\iccvPaperID{2434} 

\ificcvfinal\fi

\begin{document}

\title{DVGaze: Dual-View Gaze Estimation}

\author{Yihua Cheng\\
Beihang University\\
Beijing, China\\
{\tt\small yihua\_c@buaa.edu.cn}
\and
Feng Lu\thanks{Corresponding Author\\This work was supported by National Natural Science Foundation of China (NSFC) under grant 61972012.}\\
Beihang University\\
Beijing, China\\
{\tt\small lufeng@buaa.edu.cn }
}

\maketitle
\ificcvfinal\thispagestyle{empty}\fi

\begin{abstract}
	Gaze estimation methods estimate gaze from facial appearance with a single camera.
	However, due to the limited view of a single camera, the captured facial appearance  cannot provide complete facial information and thus complicate the gaze estimation problem. 
	Recently, camera devices are rapidly updated. 
	Dual cameras are affordable for users and have been integrated in many devices.
	This development suggests that we can further improve gaze estimation performance with dual-view gaze estimation. 
	In this paper, we propose a dual-view gaze estimation network (DV-Gaze).
	DV-Gaze estimates dual-view gaze directions from a pair of images.
	We first propose a dual-view interactive convolution (DIC) block in DV-Gaze.
	DIC blocks exchange dual-view information during convolution in multiple feature scales.
	It fuses dual-view features along epipolar lines and compensates for the original feature with the fused feature.
	We further propose a dual-view transformer to estimate gaze from dual-view features. Camera poses are encoded to indicate the position information in the transformer.
	We also consider the geometric relation between dual-view gaze directions and propose a dual-view gaze consistency loss for DV-Gaze. 
	DV-Gaze achieves state-of-the-art performance on ETH-XGaze and EVE datasets.
	Our experiments also prove the potential of dual-view gaze estimation.
	We release codes in \url{https://github.com/yihuacheng/DVGaze}.
\end{abstract}

\section{Introduction}

Human gaze provides important cues for understanding human cognition~\cite{Rahal_2019_ESP} and behavior~\cite{Dias_2020_WACV}.
It has applications in various fields such as salience detection~\cite{Wang_2019_TPAMI,Chong_2018_ECCV, Wang_2018_attentionpre}, virtual reality~\cite{patney2016towards, pai2016gazesim, mania2021gaze} and human-computer interaction~\cite{elmadjian2021gazebar,choi2022kuiper}.

Gaze estimation methods estimate human gaze from facial appearance.
Conventional gaze estimation methods usually learn person-specific eye models and fit the eye model to estimate human gaze.
These model-based methods need to build specific camera system which contains multiple IR cameras and light sources~\cite{Guestrin_2006_TBE}.
Although model-based methods have good accuracy, the complex camera system brings high costs and harms flexibility.
Appearance-based gaze estimation methods have low requirements in devices.
It only requires a single webcam to capture facial appearance and directly learns a mapping function from the appearance to gaze.
The low requirement means appearance-based gaze estimation methods have larger potential than model-based methods.
They attract much attention and become a hotspot.
However, the low requirement also brings limitations.
In particular, one single webcam has a limited field of view and therefore captures incomplete facial appearance due to facial self-occlusion.
The problem complicates gaze estimation and brings performance drop.
To handle the problem, recent methods usually design efficient feature extraction networks\cite{Cheng_2020_tip,Wang_2023_CVPR} or synthesize more images to cover data space\cite{Qin_2022_CVPR,wang2023high}. 

\begin{figure}[t]
	\begin{center}
		\includegraphics[width=\columnwidth]{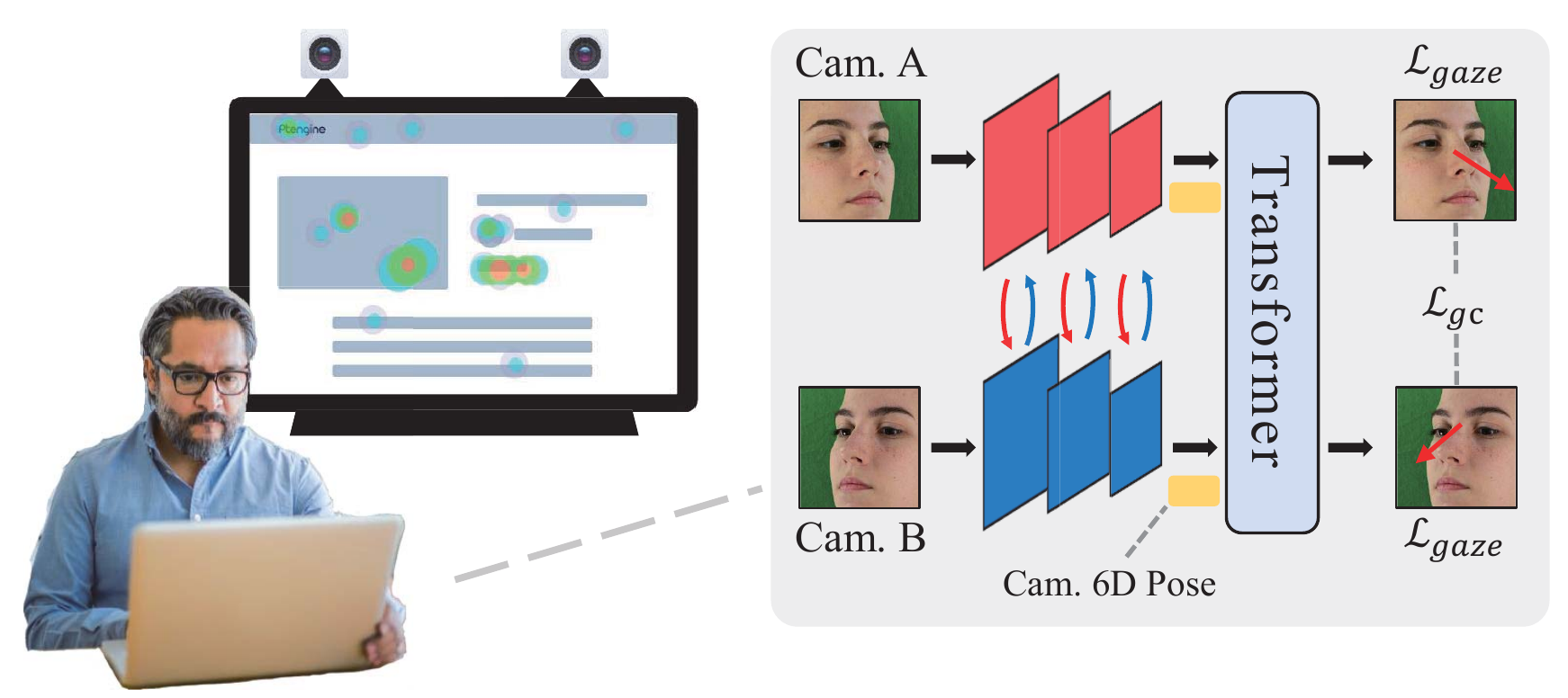}	
	\end{center}
	\caption{We explore dual-view gaze estimation in this work. 
		We propose DV-Gaze to estimate gaze directions from a pair of images.  
		DV-Gaze contains dual-view interactive convolution blocks which exchange the information of a pair of images during convolution in multiple feature scales, and a transformer to estimate gaze from dual-view feature.}
	\label{fig:overview}
\end{figure}

In this paper, we explore a new direction for gaze estimation.
Recently, camera devices are rapidly developed and the cost of camera is also decreased.
Dual cameras are affordable for users~\cite{gonzalez2019industrial} and have been applied in many devices~\cite{bestdualcamphone}.
The developing dual-camera devices make it possible and meaningful to estimate the human gaze with dual cameras.
Compared with a single camera, dual cameras provide a larger field of view.
Dual-view images can also provide more cues for gaze estimation.
These advantages indicate dual-view gaze estimation can further improve gaze estimation accuracy than conventional single-view methods.

We propose a dual-view gaze estimation network (DV-Gaze) in this work.
Common solutions usually extracts dual-view features from dual-view images and concatenate the two features for gaze estimation.
They only fuse dual-view information in a high level.
Our idea is to exchange dual-view information anywhere.
We first propose a dual-view interactive convolution (DIC) block.
DIC blocks exchange dual-view information during convolution.
The block first fuses dual-view features along epipolar lines and add the fused features back to original features for compensation.
It then performs feature extraction separately on the two compensated feature maps with convolution layers.
We stack multiple DIC blocks to exchange dual-view information in multiple feature scales during convolution.
We further propose a dual-view transformer.
The transformer estimates dual-view gaze directions from dual-view features.
We use camera pose to indicate the position information in the transformer.
We also consider the geometric relation between dual-view gaze directions and propose a dual-view gaze consistency loss function.

Overall, we summarize our contributions as following.
\begin{enumerate}[itemsep=2pt,topsep=0pt,parsep=0pt]
	\item We explore dual-view gaze estimation in this work. 
	To the best of our knowledge, our work is the first to explore dual-view gaze direction estimation. 
	\item We propose DIC blocks for dual-view feature extraction.
	The block fuses dual-view features along epipolar lines and use fused features to compensate original features. Our method exchanges dual-view feature in multiple feature scales during convolution with DIC blocks.
	\item We propose a dual-view transformer to estimate gaze from dual-view feature.
	We use camera pose to indicate the position information in the transformer.
	We also propose a dual-view gaze consistency loss which improves performance in a self-supervised manner.
\end{enumerate}

\section{Related work}

\subsection{Appearance-based Gaze Estimation}
Appearance-based gaze estimation methods aim to estimate human gaze from facial appearance~\cite{Cheng_2021_arxiv}.
These methods usually learn a mapping function $f$ from face or eye images $\bv{I}$ to human gaze $\bv{g}$, \ie, $\bv{g} = f(\bv{I})$.
Recently, deep learning shows good performance in many computer vision tasks.
Gaze estimation with deep learning also attract much attention.
Cheng~\etal propose an asymmetric regression for gaze estimation from two eye images~\cite{Cheng_2018_ECCV}.
They build an asymmetric network which adaptively assigns different weights for two-eye images.
They also further fuse facial feature to improve performance~\cite{Cheng_2020_tip}.
Park~\etal propose a pictorial representation for eye images~\cite{Park_2018_ECCV}.
They eliminate the identity difference between subjects by generating the pictorial representation and estimate gaze from the representation.
Chen~\etal use dilated convolutional network to capture the subtle changes in eye images~\cite{Chen_2019_ACCV}.
Cheng~\etal propose a coarse-to-fine network to integrate face and eye images~\cite{Cheng_2020_AAAI}.
They estimate basic gaze directions from face images and refine basic gaze directions with eye images.
These methods all achieve great performance in many benchmarks.
However, the performance of these methods still cannot meet some high accuracy requirements.
An accurate gaze estimation model is always crucial and demanded.

\subsection{Gaze Estimation with Multiple Cameras}
Gaze estimation with multiple cameras is always a hot topic in conventional gaze estimation methods.
These methods usually build multi-camera systems and estimate human gaze with geometric eye models~\cite{Guestrin_2006_TBE}.
Tonsen~\etal~embed multiple millimeter-sized RGB cameras
into a normal glasses frame. They use multi-layer perceptrons
to process the eye images captured by different cameras
and concatenate the extracted feature to estimate gaze\cite{Tonsen_2017_IMWUT}.
Common multi-camera gaze estimation systems are usually set in XR devices, such as Meta Quest Pro. 

Deep learning based gaze estimation methods attract much attention in the last decade.
However, most of methods are proposed for single-view gaze estimation.
There are merely a few methods for multi-view gaze estimation.
Lian \etal estimate points of gaze with three cameras~\cite{Lian_2019_TNNLS} . They extract feature maps of three views and use a max-pooling layer between multi-views to extract feature.
Kim \etal estimate gaze zones with three cameras~\cite{kim2020preliminary}. They simply concatenate extracted feature of three views for prediction.

\begin{figure*}[t]
	\begin{center}
		\includegraphics[width=1.95\columnwidth]{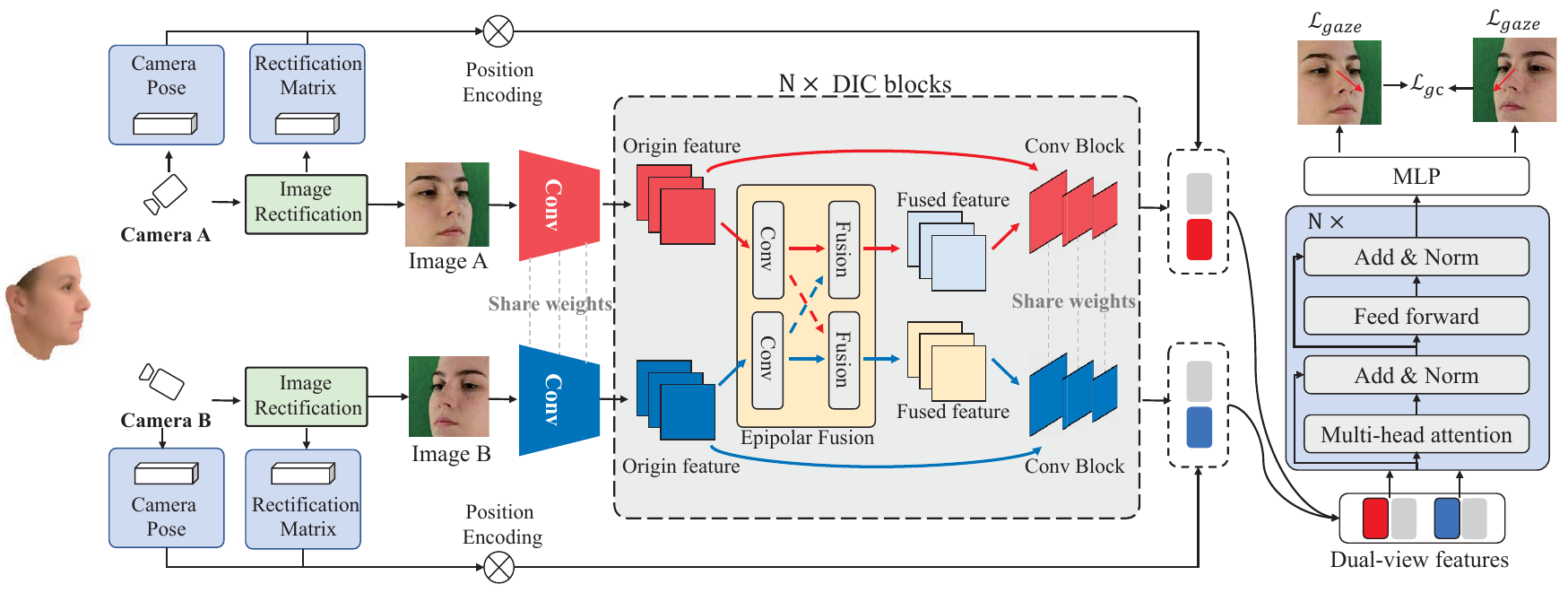}	
	\end{center}
	\caption{The pipeline of our dual-view gaze estimation. Two cameras capture dual-view images of human facial images. We first perform image rectification method on both images. We save the rectification matrix and origin camera pose for the rest steps. We input two rectified images into DV-Gaze. DV-Gaze first performs a primary convolution to extract dual-view features. We then send dual-view features into DIC blocks. The DIC block fuses dual-view feature along epipolar lines and add fused feature back to original features to mix information. We further perform feature extraction on the feature with convolution layers. We stack multiple DIC blocks to exchange dual-view information in different feature scales. We propose a dual-view transformer to estimate gaze from dual-view feature. We encode camera pose as the position feature in transformer. The loss function of DV-Gaze consists of common gaze loss and a self-supervised dual-view gaze consistency loss.  }
	\label{fig:network}
\end{figure*}

\section{Preliminaries}

\subsection{Dual-View Gaze Estimation}
We explore dual-view gaze estimation in this paper. We first define some notations and describe the task of dual-view gaze estimation.

We assume there are two cameras $\cama$ and $\camb$. In the rest of this section, we use superscripts $^\cama$ and $^\camb$ to identify different cameras.  
The two cameras capture face images $\bv{I}^\cama$ and $\bv{I}^\camb$.
Dual-view gaze estimation aims to learn a mapping function $f$ from $\bv{I}^\cama$ and $\bv{I}^\camb$ to gaze directions $\bv{g}^\cama$ and $\bv{g}^\camb$, \ie,
\begin{equation}
	(\bv{g}^\cama, \bv{g}^\camb) = f(\bv{I}^\cama, \bv{I}^\camb).
\end{equation}
The two gaze directions are separately defined in the camera coordinate systems (CCS) of dual cameras.
It is obvious $\bv{g}^\cama$ and $\bv{g}^\camb$ can be converted to each other with camera rotation.

We also make following assumptions. 1) The 6D pose of two cameras are fixed and known. The pose can be obtained by camera calibration. We use $\bv{R}_\c\in\mathbb{R}^{3\times3}$ to represent an rotation matrix and $\bv{t}_\c\in \mathbb{R}^{3\times1}$ to represent a translation vector, \ie, given a point $p$ in the world coordinate system, we can convert it into CCS of camera A as $\bv{R}_\c^\cama p + \bv{t}_\c^\cama$. 2) We have head pose $\bv{R}_\h\in\mathbb{R}^{3\times3}$ and $\bv{t}_\h\in \mathbb{R}^{3\times1}$ in each view. The head pose can be obtained by existing head trackers~\cite{lepetit_2009_epnp}.

\subsection{Dual-View Images Rectification}
\label{rectification}
A key information in dual-view system is the correspondence between two cameras.
Epipolar lines indicate the correspondence that two correctly matched points should lie on their corresponding epipolar lines.
In this work, we utilize the correspondence to fuse dual-view feature.
To simplify the process of finding epipolar lines, we use image rectification methods~\cite{Zhang_2018_etra} to make epipolar lines horizontal.

We briefly introduce the image rectification method in this section.
We set a reference point in face images, \eg, face centers.
The image rectification method computes a transformation matrix $\bv{M} = \bv{SR}_{\rec}$, where $\bv{S}$ is a scale matrix and $\bv{R}_{\rec}$ is an rotation matrix.
The scale matrix scales face images so that the distance between the reference point and virtual cameras is constant.
The rotation matrix transforms face images via perspective transformation. 
It ensures virtual cameras point at the reference point, \ie, face centers are located in image centers, and cancel the row axis of head pose.
Please refer \cite{Cheng_2021_arxiv} for more technical details.

After the image rectification, epipolar lines are horizontal and lie on corresponding rows.
Meanwhile, the methods perform perspective transformation for the rectification. This operation changes virtual camera pose of images.
The new camera pose can be computed by $\bv{R}_{\rec}\bv{R}_\c$.

\section{Dual-View Gaze Estimation}

\subsection{Overview}
In this section, we propose a dual-view gaze estimation network (DV-Gaze).
We show the pipeline of DV-Gaze in~\fref{fig:network}.
We first use a convolution layer to extract dual-view features from dual-view images.
We then propose DIC blocks.
The block fuses dual-view feature maps along epipolar lines and uses fused feature to compensate origin feature.
It performs feature extraction on the compensated feature with convolution layers.
We stack multiple DIC blocks to exchange dual-view information in multiple feature scales.
We also propose a dual-view transformer to estimate gaze from dual-view feature. We use camera pose to indicate the position information in the transformer. 
DV-Gaze predicts dual-view gaze directions from dual-view images.
We use common gaze loss,~\ie~L1 loss, as loss function.
We also propose a self-supervised dual-view gaze consistency loss based on the geometric relation of dual-view gaze.

In the rest of this section, we first introduce the DIC block and the dual-view transformer.
We then describe DV-Gaze in detail. 
We finally show all loss functions of DV-Gaze.

\begin{figure*}[t]
	\begin{center}
		\includegraphics[width=2\columnwidth]{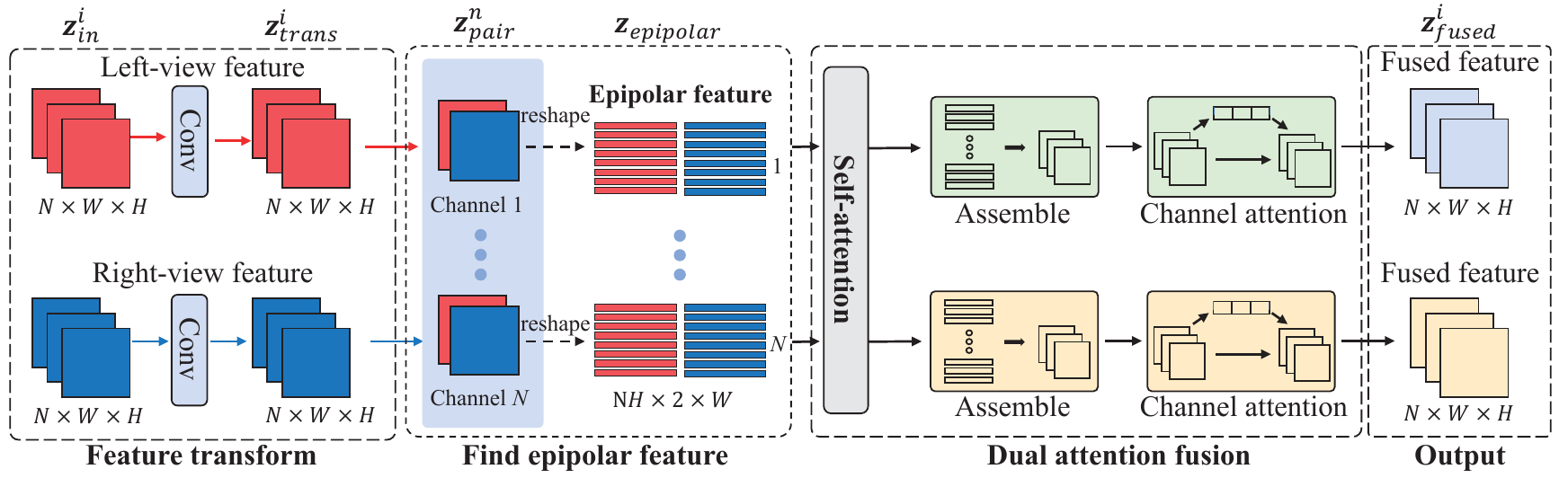}	
	\end{center}
	\caption{The pipeline of epipolar fusion. We first perform a feature transform on origin feature. To reduce computational resource, we use the bottleneck architecture of residual block for the feature transform. We then aggregate feature maps and find epipolar feature in channels. Due to the image rectification in data pre-processing, the epipolar feature is the corresponding rows in two feature maps. We use a self-attention mechanism to fuse epipolar feature and assemble feature vectors into feature maps. We also apply a channel attention in the feature maps. The output contains two feature maps which have the same size as input feature.   }
	\label{fig:epipolar}
\end{figure*}

\subsection{Dual-View Interactive Convolution Block}
The two cameras in the dual-view system have different camera positions.
This means the image of each view would miss some information about human gaze.
It is intuitive that the performance would be improved if we exchange dual-view information to compensate the missing information.

We propose DIC blocks to exchange dual-view information during convolution.
It fuses  dual-view feature and adds the fused features back to the original feature to mix information. 
Simply speaking, common convolution blocks extracts feature as 
\begin{equation}
	\bv{z}_{\mathrm{out}}^i = \mathrm{Conv}(\bv{z}_\mathrm{in}^i), i\in\{A, B\}
	\label{equ:common}
\end{equation}
where $\bv{z}_\mathrm{in}^i\in \mathbb{R}^{N\times H\times W}$ is the origin feature of each view and $\bv{z}^i_\mathrm{out}$ is the extracted feature.

DIC blocks aim to fuse dual-view features $\left\{\bv{z}_\mathrm{in}^i\right\}$ to learn dual-view fused features $\left\{\bv{z}_\mathrm{fused}^i\in \mathbb{R}^{N\times H\times W}\right\}$ to compensate $\bv{z}_\mathrm{in}^i$. The formulation of DIC block is 
\begin{equation}
	\bv{z}_\mathrm{out}^i = \mathrm{Conv}\left(\bv{z}_\mathrm{in}^i + \bv{z}_\mathrm{fused}^i\right),
	\label{equ:compensation}
\end{equation}

DIC blocks contain three steps to learn fused features.
We show the pipeline in~\fref{fig:epipolar}.

\textbf{1) Feature transform.} 
We first use two convolution blocks to respectively transform dual-view features. 
We send \{$\bv{z}_\mathrm{in}^i$\} into residual blocks~\cite{He_2016_CVPR} and get feature maps $\left\{\bv{z}_\mathrm{trans}^i\right\}$, where $\bv{z}_\mathrm{trans}^i$ has the same dimension as $\bv{z}_{in}^i$.
To reduce computational resource, we add a bottleneck in the residual block.
Overall, the residual block contains three convolution layers,
where the kernel size is set as  $3\times3$ for the second layer and $1 \times 1$ for the first and third layers. 
The number of kernel is respectively set as $\left(\frac{N}{4}, \frac{N}{4}, N\right)$.

\textbf{2) Find epipolar feature.} 
Epipolar lines indicate the correspondence between dual cameras.
DIC blocks aim to fuse dual-view features along epipolar lines.
We define a pair of features along epipolar lines as epipolar features.
In this step, our aim is to find epipolar feature from $\left\{\bv{z}_\mathrm{trans}^i\right\}$.

We aggregate feature maps $\bv{z}_\mathrm{trans}^A$ and $\bv{z}_\mathrm{trans}^B$ in channels. 
As a result, we have $N$ pairs of feature maps. We denote a pair of feature maps as $\bv{z}_\mathrm{pair}^n \in \mathbb{R}^{2\times H\times W}$, where $n\in\{1...N\}$.
The next goal is to find epipolar feature from $\bv{z}_\mathrm{pair}^n$.

As described in \sref{rectification}, we rectify images in data pre-processing.
It means epipolar lines are horizontal and lie on corresponding rows.
Therefore, we sample epipolar features from the corresponding rows of $\bv{z}_\mathrm{pair}^n$,~\ie~$\bv{z}_\mathrm{pair}^n$ contains $H$ epipolar features and the shape of each epipolar feature is $2\times W$.
We sample a total of $NH$ epipolar features from $\{\bv{z}_\mathrm{pair}^n\}$.
We define all epipolar features as $\bv{z}_\mathrm{epipolar}\in\mathbb{R}^{NH\times 2\times W}$.

\textbf{3) Dual attention fusion.} 
Our goal is to learn a mapping function $\psi:\mathbb{R}^{2\times W}\rightarrow\mathbb{R}^W$. 
The mapping function learns fused features from epipolar features. 

We use a self-attention module~\cite{vaswani2017attention} to fuse epipolar features.
We use two MLP layers to learn the query and the key from epipolar features and directly set the value as epipolar features.
The self-attention module adaptively learns weights from the query and the key. The fused feature is the weighted-sum of the value.
We reassemble all fused feature vectors and get fused feature maps $\bv{z}_\mathrm{fused}\in \mathbb{R}^{N\times H\times W}$.
Note that, the self-attention module outputs two fused feature, \ie, the self-attention is a mapping function, $\mathbb{R}^{2\times W}\rightarrow\mathbb{R}^{2\times W}$.
We respectively reassemble them and get $\bv{z}_\mathrm{fused}^\cama$ and $\bv{z}_\mathrm{fused}^\camb$.

We also add a channel attention module after the self-attention module.
We input $\bv{z}_\mathrm{fused}^A$ and $\bv{z}_\mathrm{fused}^B$ into an average pooling layer and use the output feature to learn weights for each channel.
We multiply the weights with the two fused features and get final fused features. We slightly abuse the notation here and also define the final fused feature as $\bv{z}_\mathrm{fused}^A$ and $\bv{z}_\mathrm{fused}^B$.
We use the two feature to respectively compensate original dual-view feature as \eref{equ:compensation}.

\subsection{Dual-View Transformer}
We use a transformer encoder to estimate gaze from dual-view feature.
We assume the dual-view features as $\left\{\bv{z}_\mathrm{vec}^i\right\}$.
We can formulate this module as 
\begin{equation}
	(\bv{\hat{g}}^\cama, \bv{\hat{g}}^\camb) = \mathrm{Transformer}(\bv{z}_\mathrm{vec}^\cama, \bv{z}_\mathrm{vec}^\camb),
\end{equation}
where we use $\bv{\hat{g}}^i$ to represent the predicted value.
Besides, The camera pose is an important information in the dual-view camera system.
It indicates the relations between dual-view images.
We also input camera pose into the transformer for gaze estimation.
We consider transformers usually utilize position information to identify input features, \eg, position embedding in ViT\cite{dosovitskiy2020image}.
We use the camera pose to indicate the position information in our work.

More concretely, we use camera orientation and translation vectors to represent position feature. 
As described in the preliminaries section, we have camera positions $\left\{\bv{R}_\c^i, \bv{t}_\c^i\right\}$ via calibration and the rotation matrix $\left\{\bv{R}_\mathrm{rec}^i\right\}$ in image rectification. 
Note that $\left\{\bv{R}_\mathrm{rec}^i\right\}$ is different for each image.
We multiply $\bv{R}_\rec^i$ and $\bv{R}_\c^i$ to get the rotation matrix of virtual cameras and use the third column in the rotation matrix to represent camera orientation.
\begin{equation}
	\bv{z}_\mathrm{rot}^i = \bv{R}_\rec^i\bv{R}_\c^i\left[:, 2\right]
\end{equation} 
Let $\bv{z}_{pos}$ for the position feature,  we have 
\begin{equation}
	\bv{z}_\mathrm{pos}^i= [\bv{z}_\mathrm{rot}^i; \bv{t}_\c^i],
\end{equation} 
where $[\quad]$ is a concatenation operation and $\bv{z}_\mathrm{pos}^i\in\mathbb{R}^6$.

Besides, the position features in transformers usually are discrete values.
Transformers use position embedding to expand the position feature to a large dimension~\cite{dosovitskiy2020image}.
However, $\bv{z}_\mathrm{pos}^i$ is a continuous vector and it is not applicable to use the position embedding.
In our work, we follow NeRF \cite{mildenhall_2020_nerf} and use multi-MLP layers to expand $\bv{z}_{pos}^i$.
We first expand the dimension with positional encoding which enables the position feature to contain more high frequency information. The positional encoding is 
\begin{equation}
	\gamma(p) = \left\{\sin(2^k \pi p), \cos(2^k \pi p), p \right\}_{k=0}^L,
	\label{eq:enc}
\end{equation}
where $L$ is length of the encoding.

We respectively apply the function into each dimension of $\bv{z}_\mathrm{pos}^i$. The positional encoding can also be represented as a function $\Phi:\mathbb{R}^6\rightarrow\mathbb{R}^{6*(3L)}$.
We then use a multi-MLP layers $\Pi: \mathbb{R}^{6*(3L)}\rightarrow\mathbb{R}^{d}$ to learn the final position feature, where $d$ is the length of $\bv{z}_\mathrm{vec}^i$. The whole process can be formulated as
\begin{equation}
	\bv{z}_\mathrm{pos\_exp}^{i} = \Pi(\Phi(\bv{z}_\mathrm{pos}^i))
	\label{eq:enc}
\end{equation}

We input dual-view feature vectors $\left\{\bv{z}_\mathrm{vec}^i\right\}$ with corresponding position feature $\left\{\bv{z}_\mathrm{pos\_exp}^i\right\}$ into transformers and use a MLP head to predict gaze. 

\subsection{Dual-View Gaze Estimation Network}
We propose dual-view gaze estimation network in this section.
The architecture of DV-Gaze is shown in \fref{fig:network}.
Two cameras capture human facial appearance and provide two facial images.
We first perform image rectification in the two images and input rectified dual-view images into DV-Gaze.  
Although we propose DIC block to fuse dual-view feature during convolution, it is unreasonable that we directly fuse raw dual-view images.
Therefore, we first use a convolution layer to respectively extract primary dual-view feature maps from images.
We then send the dual-view feature maps into DIC blocks for feature extraction.
We stack multiple DIC blocks and obtain the output of the final DIC block.
The outputs are dual-view feature maps.
We use an average pooling layer and a MLP layer to map the two feature maps into the dual-view feature vectors $\bv{z}_\mathrm{vec}^\cama$ and $\bv{z}_\mathrm{vec}^\camb$. 
We finally feed the dual-view feature vectors into the dual-view transformer to estimate dual-view gaze directions.

\begin{figure}[t]
	\begin{minipage}[t]{0.23\textwidth}
		\includegraphics[width=1\columnwidth]{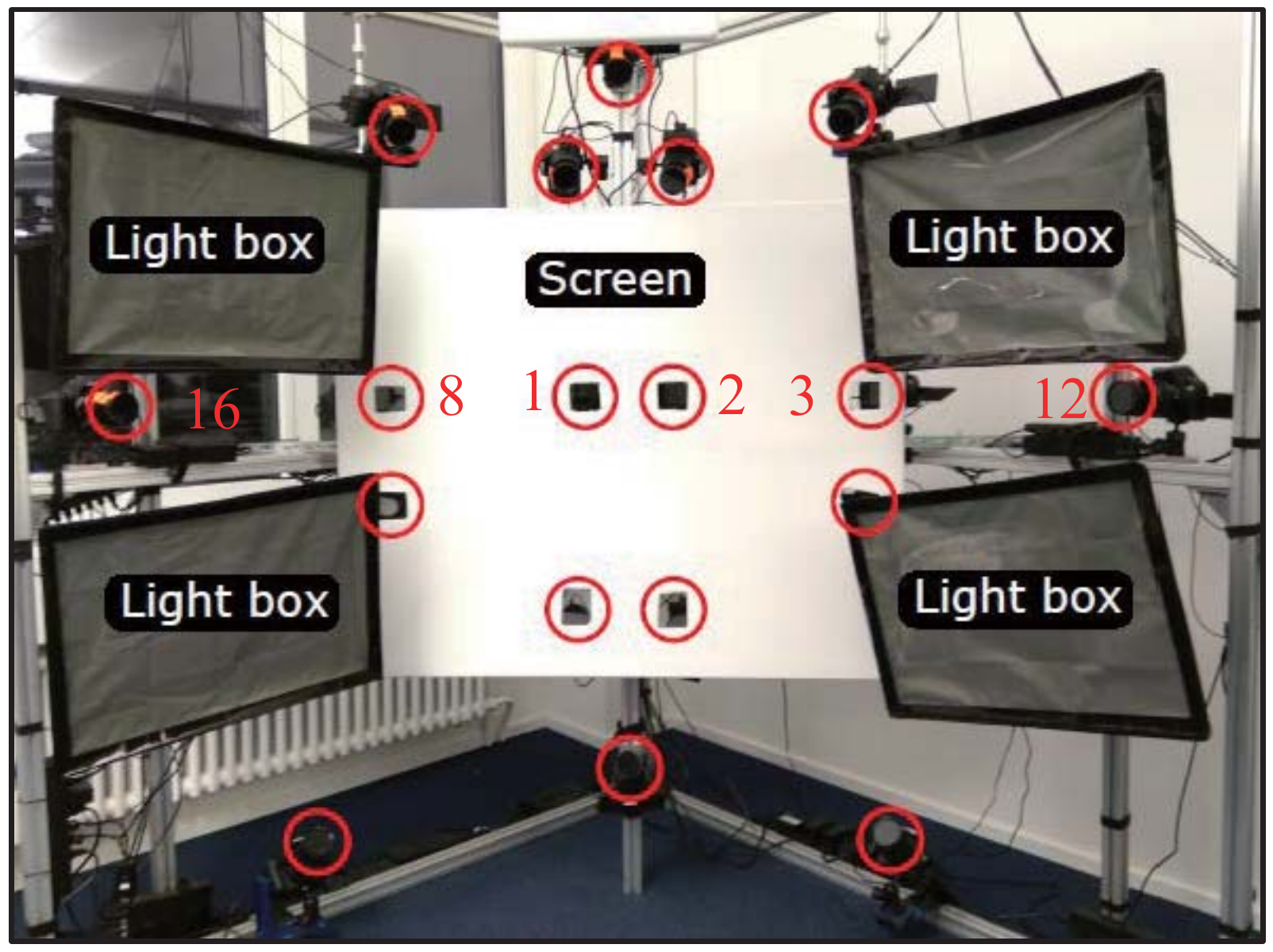}
	\end{minipage}
	\hfill
	\begin{minipage}[t]{0.23\textwidth}
		\includegraphics[width=1\columnwidth]{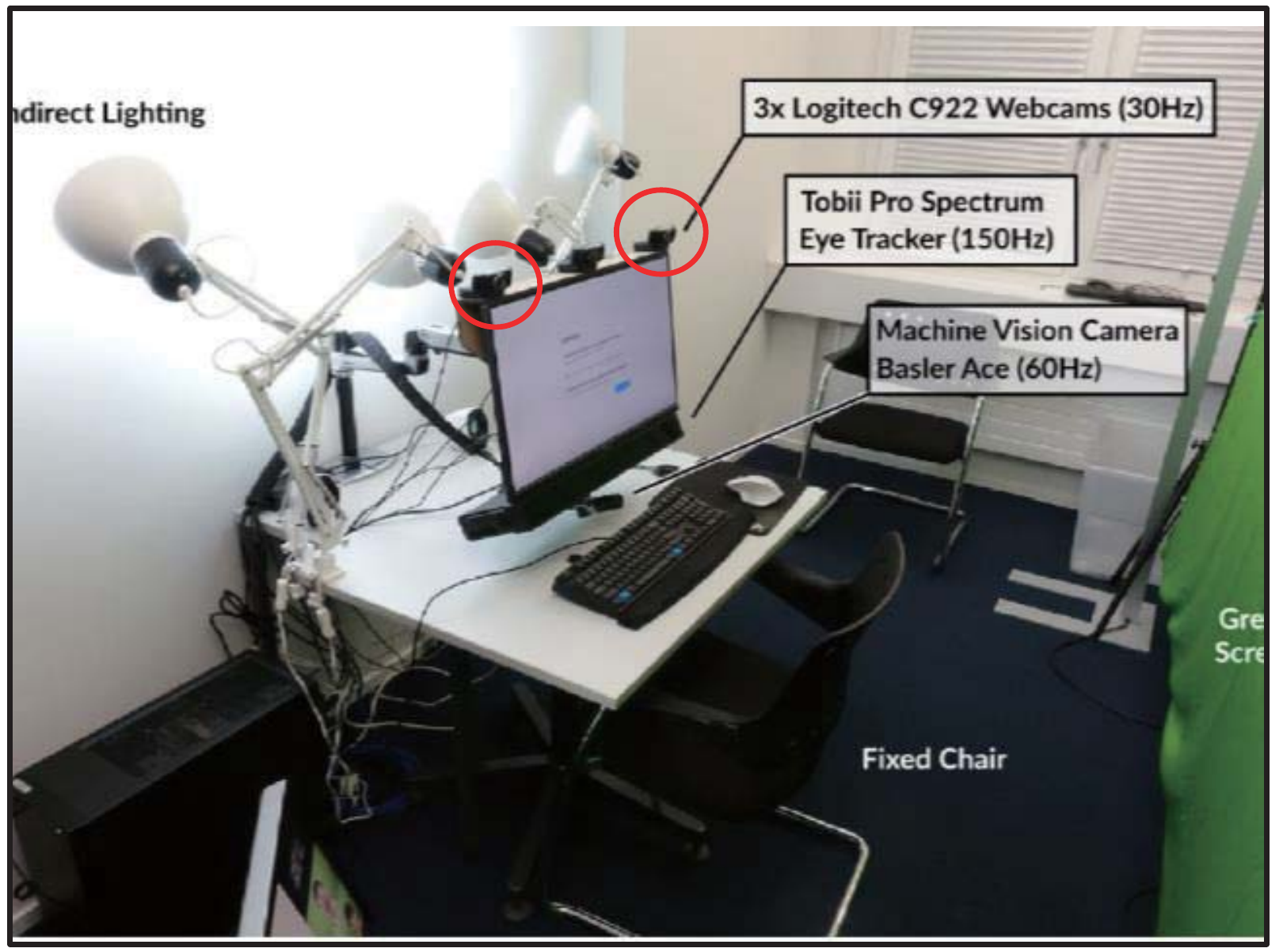}
	\end{minipage}
	\caption{We show the camera position in ETH-XGaze (left)\cite{Zhang_2020_ECCV} and EVE (right)~\cite{park_2020_eccv} dataset. 
		We use the No.3 and No.8 cameras in ETH-XGaze and the two cameras on the edge of screen in EVE.  }
	\label{fig:camera}
\end{figure}

The DV-Gaze estimates gaze $\hat{\bv{g}}^\cama$ and $\hat{\bv{g}}^\camb$ from dual-view images, where $\hat{\bv{g}}^i\in \mathbb{R}^2$ and contains pitch and yaw.
We use L1 loss to measure the loss of gaze estimation
\begin{equation}
	\mathcal{L}_{gaze} = \sum_{i\in\{A, B\}, j\in\{1...K\}}\left |\bv{g}_j^i - \hat{\bv{g}}_j^i\right|, 
\end{equation}
where $K$ denotes the number of all data.

\textbf{Dual-view gaze consistency loss.} In dual-view gaze estimation, the gaze directions of dual views can be converted into each other if camera positions are known.
Therefore, we design a self-supervised dual-view gaze consistency loss.
The loss requires dual-view gaze directions are the same when we convert them into the same coordinate system.
In detail, we convert $\{\hat{\bv{g}}^i\}$ into the world coordinate system and use MSE loss to measure the error. 
The gaze consistency loss is computed as 
\begin{equation}
	\mathcal{L}_{gc} = \left| { \left(\bv{R}_\rec^\cama \bv{R}_\c^\cama \right)}^{-1} \hat{\bv{g}}_\mathrm{3d}^\cama - {\left(\bv{R}_\rec^\camb \bv{R}_\c^\camb\right)}^{-1} \hat{\bv{g}}_\mathrm{3d}^B \right|_2, 
\end{equation}
where $\hat{\bv{g}}_\mathrm{3d}^i$ denotes gaze directions in euclidean coordinates.
$\hat{\bv{g}}_{3d}^i = (x, y, z)$ and $\hat{\bv{g}}^i = (\phi, \theta)$ can be converted as

\begin{equation}
	\begin{cases}
		x = -\cos(\phi) * \sin(\theta)\\
		y = -\sin(\phi)\\
		z = -\cos(\phi) * \sin(\theta)
	\end{cases}
\end{equation}

The final loss function is the weighted-sum of gaze loss and gaze consistency loss.
\begin{equation}
	\mathcal{L} = \alpha\mathcal{L}_{gaze} + \beta\mathcal{L}_{gc} 
\end{equation}

\section{Experiment}

\subsection{Datasets}

We conduct experiments on ETH-XGaze~\cite{Zhang_2020_ECCV} and EVE datasets~\cite{park_2020_eccv}.
The two datasets both provide large-scale multi-view images.
We show the camera layout of the two datasets in \fref{fig:camera}.
Our method estimates human gaze from dual-view cameras.
Therefore, we use the No.3 and No.8 cameras in ETH-XGaze dataset and the two cameras on the edge of screen in EVE dataset. We also denote them as left and right views in the rest experiment.

\begin{table}[t]
	\centering
	\renewcommand\arraystretch{1.1}
	\setlength\tabcolsep{7pt}
	\normalsize
	\caption{We compare our method with single-view gaze estimation methods.
		We evaluate compared methods in each view and show results in the first row.
		We also expand training set where it contains dual-view images.
		We train models in the expand set and show results in the second row. 
		We mark the \underline{highest} result among compared methods for convenience.}
	\begin{tabular}{clcccc}
		\toprule[1.5pt]
		& \multirow{2}{*}{Methods} &\multicolumn{2}{c}{ETH-XGaze} &\multicolumn{2}{c}{EVE}\\
		&& Left &Right &  Left & Right \\
		\midrule
		\multirow{4}{*}{\rotatebox{90}{Full set}} &Full Face~\cite{Zhang_2017_CVPRW} &5.11&4.77 & 4.23& 5.06\\
		&GazeTR~\cite{cheng2021gaze}  &\underline{3.46} &\underline{3.48}& 3.18&3.72\\
		&CA-Net~\cite{Cheng_2020_AAAI}&-&-&\underline{3.03}&3.81\\
		&Dilated-Net~\cite{Chen_2019_ACCV}&-&-&3.10&\underline{3.46}\\
		\midrule
		
		\multirow{4}{*}{\rotatebox{90}{Expand set}}&Full Face~\cite{Zhang_2017_CVPRW} & 4.98&4.89&  4.59&4.60\\
		&GazeTR~\cite{cheng2021gaze}  &\underline{3.11}&\underline{3.49}&\underline{2.92}&3.81 \\
		&CA-Net~\cite{Cheng_2020_AAAI}&-&-&3.31&3.87\\
		&Dilated-Net~\cite{Chen_2019_ACCV}&-&-&3.31&\underline{3.39}\\
		\midrule
		&DV-Gaze &\textbf{2.27}&\textbf{2.87}&\textbf{2.57}	&\textbf{2.68}\\
		
		\bottomrule[1.5pt]
	\end{tabular}
	\label{table:single}
\end{table}

\textbf{Data pre-processing.}
The two datasets both split origin data into training, test and validation sets.
They align face images with the rectification methods\cite{Zhang_2018_etra} and provide the rectified data. 
Therefore, we directly use the provided images in ETH-XGaze dataset for experiments.
EVE dataset provides rectified video clips.
We sample one image per fifteen frames from video clips to construct evaluation set. 

On the other hand, gaze directions are usually calculated by $\bv{g}=o_t - o_s$ in data collection, where $o_s$ represents the 3D position of gaze origin and $o_t$ represents the 3D position of gaze targets.
In the two datasets, they define face centers as the origin of gaze. 
To calculate the 3D position of face centers, they fit a 3D morphable face model and estimate 3D landmarks with the EPnP method~\cite{lepetit_2009_epnp}.
This pipeline is reasonable for single-view gaze estimation while causes inconsistent gaze between different views.
Therefore, we re-calculate gaze directions in the two datasets.
More concretely, given a pair of images, we get the corresponding 2D facial landmarks from datasets and estimate 3D landmarks using triangulation.
We select nose as gaze origin since nose is usually visible in all views. We compute new gaze directions with the new gaze origin.

Besides, the two datasets do not provide gaze targets annotation of test test.
Therefore, we also re-define the dataset splits. 
We split the origin training set into two subsets and define the two subsets as training and validation sets.
We use the original validation set as test set in our experiments.

\textbf{Evaluation metric:}
We use angular error as the evaluation metric, where a smaller error represents a better model.

\begin{table}[t]
	\centering
	\renewcommand\arraystretch{1.1}
	\setlength\tabcolsep{4pt}
	\normalsize
	\caption{We compare our method with multi-view gaze estimation methods. The column of Avg. means we average dual-view gaze directions in the world coordinate system and report the angular error between averaged gaze and ground truth. We mark the \underline{highest} performance among compared methods. }
	
	\begin{tabular}{lcccccc}
		\toprule[1.5pt]
		\multirow{2}{*}{Methods} &\multicolumn{3}{c}{ETH-XGaze} &\multicolumn{3}{c}{EVE}\\
		
		& Left &Right &Avg.\tnote{1}  & Left  & Right &Avg.\\
		\midrule
		MMGE~\cite{kim2020preliminary}  &  6.44 &7.47 &6.88 &5.71&8.19&6.19\\
		MGT~\cite{Lian_2019_TNNLS}  &2.86&3.33 &  2.97 & \underline{3.37} &3.76&3.02\\
		\midrule
		Res18 &  \underline{2.61}&\underline{3.04}& \underline{2.70}&3.39&3.58& \underline{2.97}\\
		Trans&3.41&3.15&2.74&3.76& \underline{3.50}&2.98\\
		\midrule
		DV-Gaze &\textbf{2.27}&\textbf{2.87}& \textbf{2.46} &\textbf{2.57}	&\textbf{2.68}& \textbf{2.49}\\
		\bottomrule[1.5pt]
	\end{tabular}
	
	\label{table:multi}
\end{table}
\subsection{Comparison with Single-View Methods}

Single-view gaze estimation is the main track in gaze estimation field.
Recently, many advanced methods are proposed for single-view gaze estimation.
We first conduct comparison with them to show the advantage of dual-view gaze estimation.
We select FullFace~\cite{Zhang_2017_CVPRW}, GazeTR~\cite{cheng2021gaze}, CA-Net~\cite{Cheng_2020_AAAI} and Dilated-Net~\cite{Chen_2019_ACCV} for comparison.
We evaluate these methods on ETH-XGaze and EVE datasets.
Note that, CA-Net and Dilated-Net require eye images for evaluation while ETH-XGaze cannot always provides reliable eye images due to large head pose.
We do not evaluate the two methods in ETH-XGaze dataset. 

We first evaluate these methods in each view and show the performance in the first row of \Tref{table:single}.
The result shows most of methods have $\sim 3^\circ$ accuracy in two datasets.
DV-Gaze brings large performance improvement compared with these methods.
Compared with the highest performance among compared methods, our method brings $10\%\sim 30\%$ improvement in the two datasets.
This shows the advantage of dual-view gaze estimation.

We also expand the training set of compared methods.
DV-Gaze uses dual-view images for training while compared methods only use single-view images for training.
The difference maybe bring unfair comparison.
Therefore, we expand the training set of compared method where it contains dual-view images.
We train compared methods in the expanded set and report the performance of each view in the second row of \Tref{table:single}.
Most of methods has performance improvement with the expanded set.
This is because the expanded set contains more images.
However, we notice that these improvements are all small.
The reason is they do not utilize dual-view relations. 
Our method also has the best performance in all datasets.

\begin{table}[t]
	\centering
	\renewcommand\arraystretch{1.2}
	\setlength\tabcolsep{4pt}
	\normalsize
	\caption{We conduct ablation study about the dual-view transformer and the DIC block. Note that, the ablation of DIC blocks means we do not use \eref{equ:compensation} but  \eref{equ:common} in DV-Gaze. }
	\begin{tabular}{cc|ccc|ccc}
		\toprule[1.5pt]
		
		\multirow{2}{*}{\makecell[c]{Dual-view\\ Trans.}} & \multirow{2}{*}{\makecell[c]{DIC\\Block}} &\multicolumn{3}{c|}{ETH-XGaze} &\multicolumn{3}{c}{EVE}\\
		
		&& Left &Right &Avg.& Left & Right&Avg. \\
		\midrule
		$\bv{\times}$&$\bv{\times}$&2.61&3.04& 2.70&3.39&3.58& 2.97\\
		$\checkmark$&$\bv{\times}$& 2.40 &2.99&2.51 &2.97&3.11 &2.83\\ 
		$\checkmark$&$\checkmark$ &	\textbf{2.27}&\textbf{2.87}& \textbf{2.46} &\textbf{2.57}	&\textbf{2.68}& \textbf{2.49}\\
		\bottomrule[1.5pt]
	\end{tabular}	
	\label{table:ablation}
\end{table}

\subsection{Comparison with Multi-View Methods}

We conduct comparison with multi-view methods in this section.
MGT~\cite{Lian_2019_TNNLS} estimates points of gaze from multi-view images.
The method uses eye images as input. We change the input to face images for fair comparison.
MMGE~\cite{kim2020preliminary} estimates gaze zone from multi-view images.
The model is simple and only contains three convolution blocks.
We modify the last MLP layer to estimate gaze directions.
We show their performance in \Tref{table:multi}.
Besides, dual-view gaze directions can be converted to each other via camera rotation.
We rotate dual-view gaze directions into the world coordinate system and average them to get one unique gaze direction.
We count the performance of the gaze direction and show results in the column of Avg.
The result shows DV-Gaze has better performance than both MGT and MMGE.

We also build dual-view gaze estimation networks with strong performing architecture for extensive comparison. 
We use ResNet-18~\cite{He_2016_CVPR} to extract feature from dual-view images and concatenate dual-view features to estimate dual-view gaze.
We denote the method as \textit{Res18}.
We also use ResNet-18 to extract feature and use a 6-layer transformer to fuse dual-view features\cite{cheng2021gaze}
We estimate dual-view gaze from the output of transformer.
The method is denoted as \textit{Trans}.
We show the performance of the two methods in \Tref{table:multi}.
The two methods both have good performance since they uses strong performing architecture.
However, our method also has the best performance compared with them.
The result proves our method is effective in dual-view gaze estimation.

\begin{table}[t]
	\centering
	\renewcommand\arraystretch{1.2}
	\setlength\tabcolsep{4pt}
	\normalsize
	\caption{We perform ablation study to show the effectiveness of each module. $w/o$ $\bv{z}_\mathrm{pos\_exp}$ means we do not add camera positions into dual-view transformers.
		$w/o$ $\bv{z}_\mathrm{in}$ means we do not use fused feature to compensate origin dual-view feature in \eref{equ:compensation}. $w/o$ $\mathcal{L}_{gc}$ means we remove the dual-view gaze consistency loss.}
	\begin{tabular}{l|ccc|ccc}
		\toprule[1.5pt]
		&\multicolumn{3}{c|}{ETH-XGaze} &\multicolumn{3}{c}{EVE}\\
		
		& Left &Right &Avg.& Left & Right&Avg. \\
		\midrule
		$w/o$ $\bv{z}_\mathrm{pos\_exp}$ &2.49 &3.54 & 2.70 &3.00 & 3.21 & 2.71\\
		$w/o$ $\bv{z}_\mathrm{in}$ &2.62&2.84& 2.69  &2.67 & 2.93& 2.71\\
		$w/o$ $\mathcal{L}_{gc}$&2.58 & 3.08 &2.56  &2.78 & 2.78 & 2.55\\ 
		DV-Gaze&\textbf{2.27}&\textbf{2.87}& \textbf{2.46} &\textbf{2.57}	&\textbf{2.68}& \textbf{2.49}\\
		\bottomrule[1.5pt]
	\end{tabular}	
	\label{table:ablation2}
\end{table}

\subsection{Ablation Study}
We propose DIC blocks and dual-view transformer for dual-view gaze estimation.
We conduct ablation study to demonstrate the advantage of them.
We first evaluate the basic model of DV-Gaze.
The basic model uses convolution layers of ResNet18 to extract feature from dual-view images.
It concatenates dual-view features and estimates gaze with a MLP layer.
We then add the dual-view transformer into the basic model.
The model uses dual-view transformer to estimate gaze from dual-view feature.
We also further add DIC blocks into the model, \ie, DV-Gaze.
The model uses DIC blocks to exchange dual-view information during convolution.
The performance is shown in \Tref{table:ablation}.
The result shows both the DIC block and the dual-view transformer bring sufficient performance improvement.

Besides, DV-Gaze also contains some special mechanisms.
We further evaluate these mechanisms in detail.
1) DV-Gaze adds camera position into the dual-view transformer.
We ablate the camera position and denote the experiment as $w/o~\bv{z}_\mathrm{pos\_exp}$.
2) In DIC blocks, we do not  use fused feature as output but use the fused feature to compensate origin feature,~\ie, \eref{equ:compensation}.
We also ablate $\bv{z}_\mathrm{in}$ from \eref{equ:compensation}.
3) We ablate the dual-view gaze consistency loss from loss function.
We show results of the three experiments on \Tref{table:ablation2}.
The result proves the advantage of the three mechanisms.

\subsection{Different Camera Pairs}
\begin{figure}[t]
	\includegraphics[width=1\columnwidth]{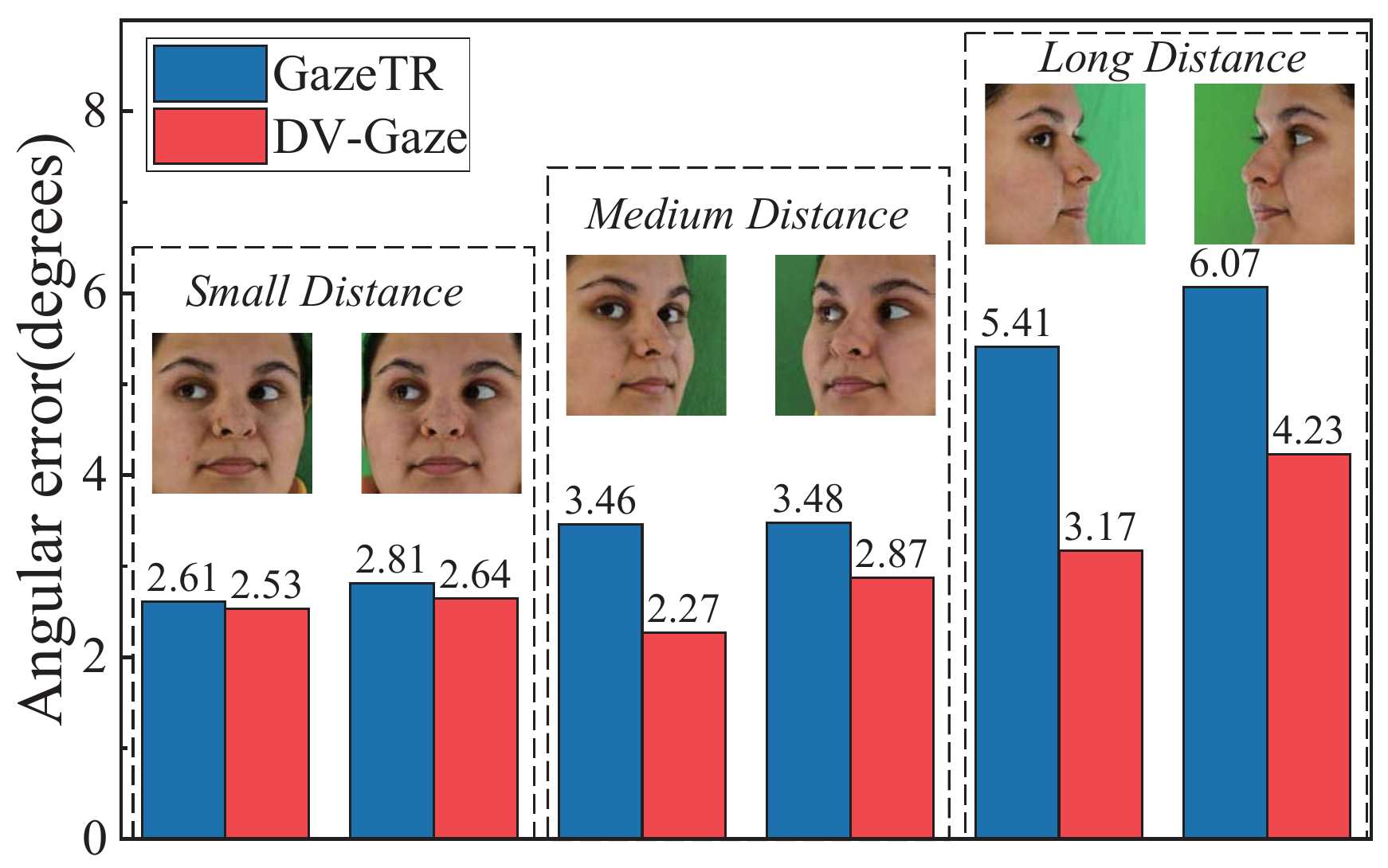}	
	\caption{We select different camera pairs in ETH-XGaze and conduct more evaluations for DV-Gaze. We also evaluate GazeTR and show the performance for comparison. It is obvious that larger camera distance brings more pixel difference between a pair of images. One eye region is not visible in the \textit{Long Distance} and DV-Gaze has the highest performance improvement in such case. On the other hand, it is interesting that camera pairs in \textit{Small distance} capture similar facial images while DV-Gaze can also bring performance improvement to some extent.   }
	\label{fig:campos}
\end{figure}

In the previous experiment, we evaluate camera pairs in ETH-XGaze and EVE dataset, where two cameras are respectively placed on the one side of a screen.
We also evaluate more camera pairs in ETH-XGaze for deep insights.
We conduct two experiments, \textit{Small Distance} and \textit{Long Distance}.
In the \textit{Small Distance}, we select No.1 camera and No.2 camera as the camera pairs.
In the \textit{Long Distance}, we select No.12 camera and No.16 camera as the camera pairs.
We also define the original camera pairs in ETH-XGaze as \textit{Medium Distance}.
We evaluate the performance of GazeTR for reference and show all results in \fref{fig:campos}. We also paste captured images of the corresponding view below the figure.

It is obvious that larger camera distance brings more vision differences in dual-view images.
Meanwhile, DV-Gaze can also bring more performance improvement.
In the \textit{Small Distance}, the captured images have no significant difference due to the small distance between camera pairs. It means the dual-camera system cannot provide extra vision information such as occlusion region than single-camera systems. However, the result shows DV-Gaze also has performance improvement in the such case despite the improvement is relatively small.
In the \textit{Medium Distance}, the eye region has a small occlusion in each images. DV-Gaze can bring $\sim 1^\circ$ performance improvement than GazeTR.
In the \textit{Long Distance}, it is obvious that one eye region is not visible in images. DV-Gaze brings large  improvement in the such case.

\subsection{Select from Image Pairs}

We propose dual-view gaze estimation in this work.
Different from conventional single-view methods, DV-Gaze fuses dual-view feature to estimate human gaze.
In this experiment, we define an oracle baseline where we select the best view from a pair of images and count the average result.

We first preform the selection on GazeTR. As shown in \Tref{table:select}, we have $2.38^\circ$ in ETH-XGaze and $2.50^\circ$ in EVE.
In fact, this result shows the upper bound of GazeTR which means we need to develop a very strong selection algorithm. 
Compared with GazeTR, DV-Gaze can easily reach the upper bound with dual-view fusion feature. DV-Gaze has $2.27^\circ$ in the right view of ETH-XGaze and $2.57^\circ$ in the left view of EVE.
This result shows the advantage of dual-view gaze estimation.
Besides, we also perform the selection based on the result of DV-Gaze. It demonstrates DV-Gaze can also improve the upper bound with dual-view feature fusion.

\begin{table}[t]
	\centering
	\renewcommand\arraystretch{1.2}
	\setlength\tabcolsep{4pt}
	\normalsize
	\caption{We obtain the results of GazeTR which is trained on each camera. We manually select the best view from each dual-view image pairs and count the average result. The result can be thought as the upper bound of the single-view method. DV-Gaze not only achieves better result than the upper bound in ETH-XGaze but also improves the upper bound in both two datasets.     }
	\begin{tabular}{l|ccc|ccc}
		\toprule[1.5pt]
		&\multicolumn{3}{c|}{ETH-XGaze} &\multicolumn{3}{c}{EVE}\\
		
		& Left &Right &Select& Left & Right&Select \\
		\midrule
		GazeTR &3.49 &3.11 & 2.38 &2.92 & 3.81 & 2.50\\
		DV-Gaze &2.87 &2.27 &2.00  &2.57 & 2.68& 2.21\\
		\bottomrule[1.5pt]
	\end{tabular}	
	\label{table:select}
\end{table}

\section{Discussion}
In this work, we have investigated the potential of dual-view gaze estimation.
We conduct experiments and prove 1) Dual-view feature fusion is necessary and useful. 2) Dual-view gaze estimation methods could achieve better performance than single-view gaze estimation methods.  
These conclusions demonstrate the advantage of dual-view gaze estimation, and prove it has more potential compared with single-view gaze estimation. 
We also show the performance of DV-Gaze in different camera pairs.
DV-Gaze brings performance improvement in all camera pairs.
It indicates dual-view gaze estimation can be useful in many scenarios, such as laptop, desktop computer, smart screen, intelligent vehicles and \etc~
On the other hand, dual-view gaze estimation also brings new challenges,~\eg, How to select the best view?
\Tref{table:select} shows we can get a better performance if we can correctly select the best view from a pair of images.
Dual-view images also provide extra stereo information. Can we utilize the information for further performance improvement?

\section{Conclusion}
We consider the limitation of single-view gaze estimation and explore dual-view gaze estimation in this work.
We propose a dual-view gaze estimation network, DV-Gaze.
DV-Gaze consists of convolutional layers, DIC blocks and a dual-view transformer.
DIC blocks fuse dual-view features along epipolar lines and compensate for original features.
We stack DIC blocks to fuse dual-view feature in multiple levels during convolution.
We feed the output of the final DIC block into the dual-view transformer.
The transformer encodes dual-view features and camera positions to estimate dual-view gaze directions.
We conduct experiments on ETH-XGaze and EVE dataset.
The experiment shows DV-Gaze has the best performance among compared methods.
Our method reveals a potential direction in gaze estimation.

\newpage

{\small
	\bibliographystyle{ieee_fullname}
	\bibliography{egbib}
}

\end{document}